\documentstyle[11pt]{article}
\newcommand{\ellipsis}{\mbox{\ldots}}

\begin{document}

\title{Minimum Description Length and Compositionality}
\author{ Wlodek Zadrozny 
}
\date{}

\maketitle

\section{Introduction}


In \cite{lp95} we have shown that the standard
definition of compositionality is formally vacuous; that is, any
semantics can be easily encoded as a compositional semantics.
We have also shown that when
compositional semantics is required to be "systematic",
it is possible to introduce a non-vacuous concept of compositionality. 
However, a technical definition of systematicity was not given in that
paper; only examples of systematic and non-systematic semantics
were presented.
As a result, although our paper clarified the
concept of compositionality, it did not solve the problem of the
systematic
assignment of meanings. In other words, we have shown that the
concept of compositionality is vacuous, but we have not replaced it
with a better definition; a definition that would both be
mathematically correct and would satisfy the common intuitions that
there are parts of grammars which seem to have compositional semantics,
and others, like idioms, that do not. We present such a non-vacuous 
definition of compositionality in this chapter.

{\it Compositionality} has been defined as the property that the
meaning of a whole is a
{\it function} of the meaning of its parts (cf. e.g. 
\cite{KeenanandFaltz85}, pp.24-25). A slightly less general definition,
e.g. \cite{Partee90}, postulates the existence of a
homomorphism from syntax to semantics.
Although intuitively clear, these definitions are not restrictive
enough. The fact that
any semantics can be encoded as a compositional
semantics has some strange consequences.
We can find, for example,
an assignment of meanings to phonems, or even the letters
of the alphabet (as the cabalists wanted), and assure
that the normal, intuitive, meaning
of any sentence is a function of the meanings of
the phonems or letters from which that sentence is composed
(cf. \cite{lp95}).\\

\noindent
To address these kind of problems we have several options. We can:\\
(a) Avoid admitting that there is a problem (e.g. by claiming
that compositionality was never intended to be expressible
in mathematical terms);\\
(b) Add additional constraints on the shape or behavior of
meaning functions (e.g.
that they are "polynomial", preserve entailment, etc.);\\
(c) Re-analyze the concept of compositionality, and the associated
intuitions. That is, that the meaning of a sentence is derived
in a systematic way from the meanings of the parts; that the meanings
of the parts have some intuitive simplicity associated with them;
and that compositionality is a gradeable property, i.e.
one way of building compositional semantics might be better than
another.

We will follow course (c). The emphasis will be on simplicity,
but the development of ideas will be formal.
(The mathematics will be relatively simple). The bottom line
will be that compositional semantics can be defined as
the simplest semantics
that obeys the compositionality principle.

\section{Basic concepts and notations}

In this section we discuss the issues in representing linguistic
information, i.e. the relationship between languages and their models. 
The first, and the simplest case to discuss is
when natural language is treated as set of words; then, 
the simplest formal model of a natural language corpus 
can be the corpus itself. A more complicated model would be a
grammar generating the sentences of the corpus; this model is 
better because it is more compact.

A more interesting case
arises when some semantics for the corpus is given. Then,
representations become less obvious, and more complicated.
Thus to keep the complexity of our presentation under control,
we will discuss
only very simple cases of natural language constructions.
This should be enough to show how to define and build compositional
semantics for small language fragments. 

Although our methods
do not depend on the size and shape of the corpora, we would like
to point out that 
computing compositional semantics for a large and real
corpus of natural language sentences would require a
separate research project, and certainly goes beyond the
the aims of this chapter.

The following issues will now be discussed: (1) representing
corpora of sentences using
grammars; (2) representing meaning functions;
(3) the size and expressive power of representations.


\subsection{Notation and essential concepts}

\subsubsection{Sentences, grammars, and meanings}

A {\it corpus} is an unordered set (bag) of sentences; a {\it sentence}
is a sequence of symbols from some alphabet.

A {\it class} is a set of sequences of
symbols from the alphabet. In our notation,
$ \{ a|b|ac \} $ denotes a class consisting of $a,b,$ and $ac$.

The {\it length}
of an expression is the number of its symbols. To make our computations
simpler, we will
assume that all symbols of the alphabet  are atomic, and hence of
length 1; same for variables.
Parentheses, commas, and most of the other notational
devices  $ \{ , \} , |, ","  \ ... $ also all have length 1;
but we will not count semicolons which we will occasionally use as
a typographical device standing for "end of line".
In several cases,
we will give the length
(in parentheses) together with an expression, e.g.
$ \{ a|b|ac \} \ \ (8)  $.

We define a (finite state) grammar {\it rule} as a sequence of classes.
E.g.  the rule
$ \{ a|b \} \{c|d\} $ describes all the combinations $ac, ad, bc, bd$.
We will go beyond finite state grammars when we discuss compositional 
semantics, and we introduce an extension of this notation then.

The reader should always remember that, {\it
mathematically},  a {\it function} is defined as a set
of pairs $ [argument, value]$.  Thus, a function does not have to
be given by a formula. A formula is not a function,
although it might define one: e.g.
a description of one entity, like energy,
depending on another, e.g. velocity, is typically given as a formula,
which defines a function (a set of pairs).

A {\it meaning function} is a
(possibly partial) function that maps sentences (and their parts)
into
(a representation of) their meanings; typically, some set-theoretic
objects like lists of features or functions.
A meaning function $\mu$  is {\it compositional} if for all elements
in its domain:
$$ \mu(s.t) =   \mu(s) \oplus \mu(t) $$
We are restricting our interest to two argument functions:
$.$ denotes the concatenation of symbols, and $\oplus$ is a function
of two arguments. However,
the same concept
can be defined if expressions are put together by other,
not necessarily binary,
operations. In literature, $\oplus$ is often taken as a composition
of functions; but in this chapter it will mostly be used as
an operator for
constructing a list, where some new attributes are added
to $\mu(s)$ and $\mu(t)$. This has the advantage of being both
conceptually simpler (no need for type raising), and closer to
the practice of computational linguistics.

\subsubsection{Minimum description length}

The {\em minimum description length (MDL) principle} was proposed by
Rissanen \cite{Rissanen82}. It states that the best theory to explain
a set of data is the one which minimizes the sum of
\begin{itemize}
\item the length, in bits, of the description of the theory, and
\item the length, in bits, of data when encoded with the help
of the theory.
\end{itemize}

In our case, the data is the language we want to describe, and the
the encoding theory is its grammar (which includes the lexicon).
The MDL principle justifies the intuition that
a more compact grammatical description is better.
At issue is what is the best encoding.
To address it, we will be simply comparing
classes of encodings.
The formal side of the argument will be kept to
the minimum; and the mathematics will
be simple --- counting 
symbols{\footnote{ We assume that the corpus contains no
errors (noise), so we do not have to worry about defining prior
distributions.}}. Counting symbols instead of bits does 
not change the line of MDL arguments, given
an alternative formulation of the MDL principle:
(p.310 of \cite{LiandVitanyi93}):

"Given a hypothesis space {\bf H},
we want to select the hypothesis $H$ such that the length of the
shortest encoding of $D$ [i.e. the data] together with the
hypothesis $H$ is minimal.
"In different applications, the hypothesis $H$ can be about different
things. For example, decision trees, finite automata, Boolean
formulas, or polynomials."\\

The important aspect of the MDL method has to do with the fact
that this complexity measure is invariant with respect
to the representation language (because of the invariance of
the Kolmogorov complexity on which it is based).
The existence
of such invariant complexity measures is
not obvious; for example, H.Simon (in \cite{Simon81}, p.228), wrote
"How complex or simple a structure is depends critically upon the
way in which we describe it. Most of the complex structures found in
the world are enormously redundant, and we can use this redundancy to
simplify their description. But to use it, to achieve this
simplification, we must find the right representation".

\subsection{Encoding a corpus of sentences}

Assume that we are given a text in an unknown language
(containing lower and uppercase letters and numbers):
$$ Xa0+Yc1+Xb0+Xc0+Ya0+Yb0$$
(We use the pluses to separate utterances, so there is no order implied.)
We are interested in building a grammar describing the text.
For a short text, the simplest grammar might in fact be the grammar
consisting of the list of all valid sentences:
$$ \{ Xa0 | Yc1 | Xb0 | Xc0 | Ya0 | Yb0 \} $$
This grammar has only 25 symbols. However, if a new corpus
is presented
$$ Za0+Wc0+Zb0+Zc0+Wa0+Wb0$$
The listing grammar would have 49 symbols, and a shorter grammar,
with only 39 symbols, could  be found:
$$
\begin{array}{lr}
\{ X | Y | Z | W \} \{ a | b \} \{ 0 \}  &  (17)   \\
\{ Y \} \{ c \} \{ 1 \} &   (9)  \\
\{ X | Z | W \} \{ c \} \{ 0 \}  & (13)
\end{array}
$$


\subsection{How to encode semantics?}

We will now examine a similar example that includes some simple
semantics.

Consider a set of nouns $n_i $, $ i \in 1..99$ and
a set of verbs
$v_j $,  $ j \in 1..9$. Let $v_0$ be {\it kick} and $n_0$ 
be {\it bucket};
and all other noun-verb combinations are intended to
have normal, "compositional" meanings.
If our corpus were to
be the $10 \times 100$ table consisting of all verb-noun
combinations:
$$  v_0 n_0 + v_1 n_0 + \ ... \ + v_j n_i  + ...  $$
we could quickly use the previous example to write a simple finite
state grammar that describes the corpus:
$$
\begin{array}{lr}
\{ v_0 | v_1 | ...  \} \{ n_0 | n_1 ...  \}  & (21 + 201)  \\
\end{array}
$$
But in this subsection we are supposed to introduce some semantics.
Thus, let our corpus consist of all those 1,000
sentences together with
their meanings, which, to keep things as simple as possible,
will be simplified to two attributes.
Also, for the reason of simplicity, we assume that only "kick bucket"
has an idiomatic meaning, and all other entries are assigned the
meaning consisting of the two attribute expression
$ [[action, v_j ], \ [object, n_i ]] $.
Hence, our corpus will look as follows: \\
\[
\begin{array}{l}
kick \ bucket \  action \  die  \  object \  nil   \\
v_1  \ bucket \  action \  v_1  \  object \  bucket \\
\mbox{} ... \\
v_j  \ n_i \  action \  v_j  \  object \  n_i \\
\mbox{} ...
\end{array}
\]
Now, notice that this corpus cannot be encoded by means of a short
finite state grammar, because of the dependence of the meanings
(i.e. the pair $ [ action,  ... , \ object, ... ] $)
on the first two elements of each sentence. We will have to
extend our grammar formalism to address this dependence
(Section 3).

\subsection{On meaning functions}

Even though we cannot encode the corpus by a short, finite state
grammar, we can easily provide for it a compositional semantics.
To avoid the complications of type raising, we
will build a homomorphic mapping from syntax to semantics. To do it,
it is enough to build meaning functions in a manner ensuring
that the
meaning of each $ v_j n_i $ is composed from the meaning of
$v_j$ and the meaning of $n_i$. Since our corpus is simple,
these meaning functions are simple, too:
For the verbs the meaning function is given by the table:
$$ [ v_0 , [verb , v_0]]  ; \  [ v_1, [verb , v_1]]
\ ... \  [ v_9, [verb , v_9]]  \ \  (90)  $$
For the nouns:
$$ \ \  [ n_0 , [noun , n_0]]; \ [ n_{1}, [noun , n_{1}]]
\ ... \ [ n_{99}, [noun , n_{99}]] \ \  (900) $$

We have represented both meaning functions as tables of symbols.
Since this chapter deals with sizes of objects, we compute them for
the meaning functions:
the size of the first function is $90 = 10 \times 9$,
and for the second one it is $900 = 100 \times 9$.
Therefore, the meaning function for the whole corpus
could be represented as a table
with 1,000 entries: \\
$$
\begin{array}{l}
\begin{array}{l}
[ [ [ verb, v_9 ] , [ noun , n_{99} ] ] ,
[ [action, v_9 ] ,  [object, n_{99} ] ] ]  \\
\ellipsis
\end{array} \\
\begin{array}{l}
[ [ [ verb, v_{j} ] , [ noun , n_{i} ] ] ,
[[action, v_j ] , [object, n_i ]]]  \\
\ellipsis
\end{array} \\
\begin{array}{l}
[[ [verb, v_1 ] , [ noun , bucket ] ] ,
[ [ action, v_{1} ], \ [object, bucket ]]]
\end{array} \\
\begin{array}{l}
[[ [verb , kick ] , [ noun ,  bucket ] ] ,
[[action, die ], [object, nil ] ]
\end{array}
\end{array}
$$
and the size of this table is $29 \times 1000$.
Finally,
the total size of the tables that describe the compositional
interpretation of the corpus is $29000+900+90$, i.e. roughly
$30,000$.
Notice that if we had more verbs and nouns, the
tables describing the meaning functions would be even
larger.\footnote{
%
The reader familiar with
\cite{lp95} should notice that the meaning functions obtained
by the solution lemma also consist of tables of element-value
pairs. It is easy to see that for the corpus we are encoding
the solution lemma produces the same meaning functions.

In the other direction, the method for deriving compositional
semantics using the minimum description length principle
(Sections 3 and 4) are directly applicable to meaning functions
obtained by the solution lemma
in \cite{lp95}, provided they are finite (which covers
the practically interesting cases); and it seems applicable
to the infinite case, if it has a finite representation.
However, we will not pursue this
connection any further.}
Also, note that we have not counted the cost of encoding
the positions of elements of the table, which would be the
$log$ of the total number of symbols in the table. This simplifying
assumption does not change anything in the strength of our
arguments (as larger tables have longer encodings).


\section{Compositional semantics through the Minimum Description
Length principle}

In this section we first extend our notation to deal with
semantic grammars. Then we apply the minimum description length
principle to construct a compact representation of our example
corpus. This experience will motivate
our new, non-vacuous definition of the notion of {\it compositional 
semantics} given in Section 4.

\subsection{Representations}

We have seen that it is impossible to efficiently encode
our semantic corpus using a finite state grammar.
Therefore, we have to make our representation of grammars more
expressive (at the price of a slightly bigger interpreter). Namely,
we will allow a simple form of unification. \\

\noindent
{\bf Example}.
Assume we do not want
$ \{ a | b \} \{ a | b | d   \}  $  to generate
$ab$. We can do it by changing the notation:
$$
\begin{array}{l}
X = \{ a | b  \}  \\
\{ X \} \{ X | d   \}
\end{array}
$$
The intention is simple: first, we define a class variable ($X$)
for the class consisting of elements $a$ and $b$; then, we
generate all strings using the rule with
variable $X$: $XX$ and $Xd$; and finally we
substitute for $X$ all its possible values, which produces
$aa$, $ad$, $bb$ and $bd$.\\

More generally, let us assume that we have an alphabet
$a_{1}, a_{2}, ... $, and a set of (class) variables
$X_{1}, X_{2}, ... $. A {\it grammar term}, denoted by $t_i$,
is either a sequence of symbols from the alphabet or a class variable.
By a {\it grammar rule} we will understand one of the
three expressions
$$
\begin{array}{l}
X_m = \{ X_j \} \ ... \ \{ X_n \} \\
X_m = \{ t_{i} | ... | t_{k} \}  \\
X_m = X_l
\end{array}
$$
A {\it grammar} is a collection of grammar rules.
The language generated by the grammar is defined as above.

Thus, new classes are obtained from elements of the alphabet
by either
the {\it merge} operation, which on two classes $X$ and $Y$ produces
a new class $C_{XY}$  consisting of the set theoretic union
of the two:
$ C_{XY} =  \{ X  | Y \} $; or by concatenating elements of
two or more classes. We permit renaming of classes, because
we want to be able to express constructions like
$ Noun_{person} \ know \ Noun_{person} $:
$$
\begin{array}{l}
N1 = Noun_{person} \\
N2 = Noun_{person} \\
\{N1\} \{know\} \{N2\}
\end{array}
$$


\subsection{An MDL algorithm for encoding semantic corpora}

In \cite{Grunwald96} a greedy algorithm for clustering elements
into classes  is presented. The algorithm is 
trying{\footnote{There is no guarantee 
that the algorithm will produce
the minimum length description.}}
to
minimize the
description length of grammars according to the MDL principle.
This algorithm would not work properly on our semantic
corpus, because Grunwald's
representation language is not expressive enough.
However, the representation of grammars we introduced above
allows us to use the same algorithm with only minor changes. \\

The basic steps of the greedy MDL algorithm are as follows:
\begin{enumerate}
\item Assign separate class $\{w\}$ to each different word (symbol)
in the corpus.
Substitute the class for each word in the corpus.
This is the initial grammar $G$.
\item Compute the total description length (DL) of the corpus.
(I.e. the sum of the DL of the corpus given $G$ and the DL of $G$).
\item Compute for all pairs of classes in $G$ the difference in
in DL that would result from a merge of these two classes.
\item Compute for all pairs of classes $ C_i , C_j $
in $G$ the difference in
in DL that would result from a construction of a new class
given by the concatenation rules
$$ X = \{ C_i \} \{ C_j  \} $$
\item If there is one or more operations that result in a smaller
new DL, perform the operation that produces the smallest new DL,
and go to Step 2.
\item Else Stop
\end{enumerate}


\subsection{Applying the MDL algorithm to encode
a semantic corpus}

We will now show how the algorithm applies to our corpus of
1,000 sentences. By {\bf Step 1},
the initial grammar $G_0 $ looks as follows: \\

{\bf Initial grammar} $G_0 $:
$$
\begin{array}{l}
\{ kick \} \ \{ bucket \} \
\{ action \} \{ die \} \{ object \} \{ nil \}  \\
\{ v_1 \} \{ bucket \}
\{ action \} \{ v_1 \} \{ object\} \{ bucket \}   \\
\mbox{}  ... \\
\{ v_j \} \{ n_i \} \{action\} \{ v_j \} \{ object\} \{ n_i \}  \\
\mbox{} ...
\end{array}
$$

{\bf Step 2}. Computing the total length: The grammar describes
the corpus. The size is
of the initial grammar is
18,000 symbols (not counting the encoding of the positions of
beginnings of each rule). For all
the grammars obtained by the steps of the algorithm,
the total length will be the size of the
grammar plus the size of the machine that generates languages
from grammars. But, since the size of this machine is constant, we can 
remove it from our considerations.\\

{\bf Step 3}. Merging.
Consider the merge operation for two nouns, and
the new class  $ N_{kl} = \{ n_k | n_l \} \  (7)  $, $ k, l > 0 $.
The resulting new description
of the corpus is shorter since it removes
20 entries with $ n_k ,  n_l  $ of total length 360,
and  adds two entries of total length 25
$$
\begin{array}{l}
\{ v_i \} \{ N_{kl} \}
\{action\} \{ v_i \} \{ object\} \{ N_{kl} \} \\
\  N_{kl} = \{ n_k | n_l \}
\end{array}
$$\\
However, the merge operation for two verbs produces a better grammar.
The new class  $ V_{kl} = \{ v_k | v_l \} \  (7)  $, $ k, l > 0 $.
removes
200 entries with $ v_k ,  v_l  $ of total length 3600,
and  adds one entry of length 25
$$
\begin{array}{l}
\{ V_{kl} \} \{ n_j \}
\{action\} \{ V_{kl} \} \{ object\} \{ n_j \} \\
V_{kl} = \{ v_k | v_l \}
\end{array}
$$\\
Notice that merging another
verb with {\it kick} would save only 199 rules, so it will
not be done in the initial stages of the application of the
algorithm.\\

{\bf Step 4}. The reader may check that this step would not reduce
the size of the grammar. (This is due to the corpus being so simple,
and without substructures worth encoding). \\


{\bf Step 5}.
The
successive merges of $v_i$'s ($ i > 0$ )
will produce the following grammar:\\

{\bf Grammar} $G_{V(1)} $:
$$
\begin{array}{lr}
V(1) = \{ v_1 | \ ... \ | v_9 \}  &   (21)   \\
\{ V(1)  \}  \{ n_0 \}  \{  action \}  \{  V(1)  \}
\{  object \} \{ n_0 \}  & (18) \\
\{ V(1)  \}  \{ n_1 \}  \{  action \}  \{  V(1)  \}
\{  object \} \{ n_1 \}  & (18) \\
\ ... & (18) \\
\{ V(1)  \}  \{ n_{99} \}  \{  action \}  \{  V(1)  \}
\{  object \} \{ n_{99} \}  & (18) \\
\{ v_0  \}  \{ n_0 \}  \{  action \}  \{  v_0  \}
\{  object \}  \{ nil \}  & (18) \\
\{ v_0  \}  \{ n_1 \}  \{  action \}  \{  v_0  \}  \{  object \}
\{  n_1 \} \ \ \   & (18) \\
\ ... &  (18) \\
\{ v_0  \}  \{ n_{99} \}  \{  action \}  \{  v_0  \}  \{  object \}
\{  n_{99} \} \       & (18)
\end{array}
$$

What happens next depends on whether our algorithm is very greedy;
namely, whether we insist that all instances of the merging
classes are replaced by the result of the merge.
If that is the case, we cannot do
the merge  $V(0) = \{ V(1) | v_0 \} $, and we will do the merge of
the nouns. These merges will produce \\

{\bf Grammar} $G_{V(1)N(1)} $:
$$
\begin{array}{lr}
V(1) = \{ v_1 | \ ... \ | v_9 \}  &   (21)   \\
N(1) = \{ n_1 | \ ... \ | n_{99} \}  &   (201)   \\
\{ V(1)  \}  \{ n_0 \}  \{  action \}  \{  V(1)  \}
\{  object \} \{ n_0 \}  & (18) \\
\{ V(1)  \}  \{ N(1) \}  \{  action \}  \{  V(1)  \}
\{  object \} \{ N(1) \}  & (18) \\
\{ v_0  \}  \{ n_0 \}  \{  action \}  \{  v_0  \}
\{  object \}  \{ nil \}  & (18) \\
\{ v_0  \}  \{ N(1)\}  \{  action \}  \{  v_0  \}  \{  object \}
\{  N(1)\} \ \ \   & (18)
\end{array}
$$

This is our final grammar ({\bf Step 6}) (if the algorithm is very
greedy). We can see that it is much
smaller than the original grammar --- its total length is less than
300 symbols (vs. 18,000); but it assumes an existence of a language 
generator. Interestingly, the grammar resembles the compositional
semantics, as usually given. The rule with $V(1)$ and $N(1)$ describes
the compositional part of the corpus;
the rule with $v_0$ and $n_0$ -- the idiomatic; other rules are
in between.\\

\subsection{Variations on the MDL algorithm}

A similar result is obtained when we do not insist that all instances
of merging classes are replaced by the result of the merge.
Starting with the grammar\\

{\bf Grammar} $G_{V(1)} $:
$$
\begin{array}{lr}
V(1) = \{ v_1 | \ ... \ | v_9 \}  &   (21)   \\
\{ V(1)  \}  \{ n_0 \}  \{  action \}  \{  V(1)  \}
\{  object \} \{ n_0 \}  & (18) \\
\{ V(1)  \}  \{ n_1 \}  \{  action \}  \{  V(1)  \}
\{  object \} \{ n_1 \}  & (18) \\
\ ... \\
\{ V(1)  \}  \{ n_{99} \}  \{  action \}  \{  V(1)  \}
\{  object \} \{ n_{99} \}  & (18) \\
\{ v_0  \}  \{ n_0 \}  \{  action \}  \{  v_0  \}
\{  object \}  \{ nil \}  & (18) \\
\{ v_0  \}  \{ n_1 \}  \{  action \}  \{  v_0  \}  \{  object \}
\{  n_1 \} \ \ \   & (18) \\
\ ... &  (18) \\
\{ v_0  \}  \{ n_{99} \}  \{  action \}  \{  v_0  \}  \{  object \}
\{  n_{99} \} \ \ \   &  (18)
\end{array}
$$\\
We can see that the merge
$V(0) = \{ v_0 | V(1) \} $ will decrease the size of the grammar
by 99 rules and result in:\\

{\bf Grammar} $G_{V(0)} $:
$$
\begin{array}{lr}
V(1) = \{ v_1 | \ ... \ | v_9 \}  &   (21)   \\
V(0) = \{ v_0 | V(1) \}  & (7) \\
\{ V(1)  \}  \{ n_0 \}  \{  action \}  \{  V(1)  \}
\{  object \} \{ n_0 \}  & (18) \\
\{ V(0)  \}  \{ n_1 \}  \{  action \}  \{  V(0)  \}
\{  object \} \{ n_1 \}  & (18) \\
\ ... \\
\{ V(0)  \}  \{ n_{99} \}  \{  action \}  \{  V(0)  \}
\{  object \} \{ n_{99} \}  & (18) \\
\{ v_0  \}  \{ n_0 \}  \{  action \}  \{  v_0  \}
\{  object \}  \{ nil \}  & (18)
\end{array}
$$\\

\noindent
The successive merging of nouns will then produce\\

{\bf Grammar} $G_{V(0)N(1)} $:
$$
\begin{array}{lr}
V(0) = \{ v_0 | V(1) \}  & (7) \\
V(1) = \{ v_1 | \ ... \ | v_9 \}  & (21)    \\
N(1) = \{ n_1 | \ ... \ | n_{99} \}  &  (201)  \\
\{ V(0) \} \{N(1) \} \{ action \} \{ V(0)  \} \{ object \}
\{ N(1) \} & (18) \\
\{ V(1)  \} \{n_0 \} \{ action \} \{ V(1)  \}
\{ object \} \{ n_0 \} & (18)  \\
\{ v_0  \} \{n_0 \} \{ action \} \{ v_0  \}
\{ object \} \{ nil \}  & (18)
\end{array}
$$ \\

If, however we do not do the
$ V(0) = \{ v_0 | V(1) \} $ merge, and proceed with the
merging of the nouns (e.g. if there were reasons to modify the
algorithm),
we get:\\

{\bf Grammar} $G_{V(1)N(0)} $:
$$
\begin{array}{lr}
V(1) = \{ v_1 | \ ... \ | v_9 \}  & (21)    \\
N(1) = \{ n_1 | \ ... \ | n_{99} \}  &  (201)  \\
N(0) = \{ n_0 | N(1)  \}  &  (7)  \\
\{ V(1)  \} \{N(0) \} \{ action \} \{ V(1)  \}
\{ object \} \{ N(0) \} & (18)  \\
\{ v_0  \} \{N(1)\} \{ action \} \{ v_0  \}
\{ object \} \{ N(1)\}  & (18) \\
\{ v_0  \} \{ n_0 \} \{ action \} \{ v_0 \} \{ object \} \{ nil \} & (18)
\end{array}
$$ \\

Finally,
if we allow some overgeneralization, we can replace the above grammars
with an even shorter grammar:\\

{\bf Grammar} $G_{V(0)N(0)} $:
$$
\begin{array}{lr}
V = \{ v_0 | \ ... \ | v_9 \}  & (23)    \\
N = \{ n_0 | \ ... \ | n_{99} \}  &  (203)  \\
\{ V  \} \{N \} \{ action \} \{ V  \}
\{ object \} \{ N \} & (18)  \\
\{ v_0  \} \{ n_0 \} \{ action \} \{ v_0 \} \{ object \} \{ nil \} & (18)
\end{array}
$$ \\
Here, clearly $v_0$ is the idiomatic element. However, both idiomatic
and non-idiomatic reading of {\em kick bucket} is allowed.
(In the previously defined grammars, we can also see the distinction 
between the idiomatic and non-idiomatic elements).


\section{A non-vacuous definition of compositionality}

The fact that that the MDL principle can produce an object
resembling a compositional semantics is crucial. It allows us
to argue for a non-vacuous definition of compositionality.

Assume that we have a corpus $S$ of sentences and their parts,
given either as a set
or generated by a grammar. Let sentences and their parts
be collections of symbols
put together by some operations; in the simplest and most important
case, by concatenation $"."$.\\

\noindent
{\bf Definition}. A meaning function $\mu$ is a
{\it compositional semantics} for the set $S$ if its domain is
contained in $S$, and  \\
{\bf a}. it satisfies the postulate of compositionality: for all
$s, \ t$ in its domain:
$$ \mu(s.t) =   \mu(s) \oplus \mu(t) $$
{\bf b}.
it is the shortest, in the sense of the Minimum Description Length 
principle, such an encoding.\\
{\bf c}. it is maximal, i.e. there is no $\mu '$ with a larger domain
that satisfies {\bf a} and {\bf b}. \\

To see better what this definition entails, let us consider our
semantic corpus again. The set $S$ consists of the 10 verbs and 100
nouns and all noun-verb combinations. The compositional function
$\mu$ assigns to
each word its category
e.g. $[n_{17}, \ noun]$. The question is how to define the
operator $\oplus$. Because of the idiom, it cannot be a total
function; hence we have to exclude from the domain of $\oplus$ the pair
$ [[ v_0 , \ verb ] , \  [ n_0 , \ noun ]] $.
The shortest description of $\oplus$ can be given
by translating the grammar of Section 3.2. First, map non-idiomatic
verbs and nouns into pairs
$ \mu( v_i ) = [ v_i , \ verb_{nonid} ] $,
$ \mu( n_j ) =  [ n_j , \ noun_{nonid} ] $, $i,j > 0$.
Then, put
$$ \oplus( [[ v , \ verb_{nonid} ] , \ [ n , \ noun_{nonid} ]] )  =
[ action . v  , \ object.n] $$
Thus defined $\mu$ and $\oplus$ correspond to the grammar
obtained by the algorithm of Section 3.2 and to the tables of
Section 2. This correspondence is not exact, because
functions $\mu$ and $\oplus$ encode only the systematic,
compositional part of the corpus. (But please note this clear
distinction between the idiomatic and the compositional parts
of the lexicon and the corpus).

However this description of the two functions is not maximal.
We obtain the maximal compositional semantics for $S$ 
by extending the above defined mapping
to all nouns
$ \mu( n_j ) =  [ n_j , \ noun ] $, $j \geq 0$, and extending the
domain of $\oplus$
$$ \oplus( [[ v , \ verb_{nonid} ] , \ [ n , \ noun ]] )  =
[ action . v  , \ object.n] $$
It is easily checked that
this is the shortest
(in the sense of the MDL) and maximal assignment of
meaning to the elements of set $S$.\footnote{We are assuming that
we have to assign the noun and verb categories to the lexical
symbols of the corpus.} Please compare this mapping 
with $G_{V(1)N(0)}$, and also note that now we have a formal  
basis for saying that (for this corpus)
it is the verb {\em kick}, and not the noun {\em bucket},
that is idiomatic.\\

What are the advantages of defining compositionality using the
Minimum Description Length principle?
1. It brings us back to the original definition
of compositionality, but makes it non-vacuous.
2. It encodes the postulate that the meaning
functions should be simple. 3. It allows us to distinguish between 
compositional and non-compositional semantics by means of systematicity,
i.e. the minimality of encodings,
as e.g. Hirst \cite{Hirst87} wanted. 4. It does not make a reference
to non-intrinsic
properties of meaning functions (like being a polynomial).
5. It works for different models of language understanding:
pipeline (syntax, semantics, pragmatics), construction grammars
(cf. \cite{Fillmal88}),
and even semantic grammars.
6. It allows us to compare different meaning functions with respect to
how compositional they are --- we can measure the size of their
domains and the length of the encodings.
Finally, this definition
might even satisfy those philosophers of language who regard
compositionality not as a formal property but
as an unattainable ideal worth striving for. This hope is based on the fact
that, given an appropriately rich model of language,
its minimum description length is, in general,
non-computable, and can only be approximated 
but never exactly computed.

\section{Discussion and Conclusions}

\noindent
{\bf Lambdas, approximations, and the minimum description length}

Assuming that we have a $\lambda $-expressions interpreter
(e.g. a lisp program),
we could describe the meaning functions of Section 3 as:
$$
\begin{array}{l}
\lambda X . [noun, X] \\
\lambda Y . [verb, Y]   \\
\lambda [verb,Y] [noun, X] . [[action, Y],[object,X]] \\
\lambda [verb,kick] [noun, bucket] . [[action, die],[object,nil]] \\
\end{array}
$$
The approximate
total size of this description is $size(\lambda - interpreter)$
+ 66 (the above definitions) +  110 (to describe
the domains of the first two functions).

Clearly, the last lambda expression
corresponds to an idiomatic meaning. But,
note that this definition assigns also the non-idiomatic meaning
to "kick bucket". Thus, although much simpler, it does not
exactly correspond to the original meaning function.
It does however correspond to grammar $G_{V(0)N(0)}$ of the
previous section.
Also,
representations that ignore exceptions are more often found
in the literature. This point may be worth pursuing: Savitch in
\cite{Savitch93} argues that approximate representation in a more
expressive language can be
more compact. For approximate representations that overgeneralize,
the idiomaticity of an expression can be defined as
the existence of a more specific definition of its meaning. \\

\noindent
{\bf Bridging linguistic and probabilistic approaches to natural
language}

The relationship between linguistics principles
and the MDL method is not completely surprising.
We used the MDL principle
in \cite{consmath1}  to argue for a construction-based approach
to language understanding (cf. \cite{Fillmal88}). After setting up a
formal model based on linguistic and computational evidence, 
we applied the MDL principle to prove
that construction-based representations are
at least an order of magnitude more compact that the corresponding 
lexicalized representations of the same linguistic data.
The argument presented there suggests that in 
building compositional semantics
we might be better off when the language is build by means of reach 
combinatorics (constructions), than by the concatenation of lexical
items. However, this hypothesis remains to be proved.\\

It is known that the most important rules of statistical reasoning,
the maximum likelihood method, the maximum entropy method,
the Bayes rule and the minimum description length,
are all closely related
(cf. pp. 275-321 of \cite{LiandVitanyi93}). From the material of
Sections 3 and 4 we can
see that compositionality is closely related to the MDL principle;
thus, it is possible to imagine bringing together linguistic and 
statistical methods for natural language understanding. For example,
starting with semantic classes of \cite{Dixon91} continue derivation
of semantic model for a large corpus using the method of Section 3
with the computational implementation along the lines of 
\cite{Brownetal92}. \\

\noindent
{\bf Conclusion}

We have redefined the linguistic
concept of compositionality as the simplest maximal description of
data that satisfies the postulate that the meaning of the whole
is a function of the meaning of its parts.
By justifying compositionality by the minimum description length 
principle, we have placed the intuitive idea that the meaning of a sentence 
is a combination of the meanings of its constituents on a firm mathematical 
foundation. 

This new, non-vacuous definition of compositionality 
is intuitive and allows us to distinguish between
compositional and non-compositional semantics, and between
idiomatic and non-idiomatic expressions. It is not ad hoc,
since it does not make any references to non-intrinsic
properties of meaning functions (like being a polynomial).
It works for different models of language understanding. Moreover,
it allows us to compare different meaning functions with respect to
how compositional they are.

Finally, because of the close relationship between
the minimum description length principle and probability, 
the approach proposed in this chapter 
should bridge logic-based and 
statistics-based approaches to language understanding.

\end{document}